\documentclass[conference]{IEEEtran}
\IEEEoverridecommandlockouts
\usepackage{cite}
\usepackage{amsmath,amssymb,amsfonts}
\usepackage{graphicx}
\usepackage{textcomp}
\usepackage{xcolor}
\usepackage{algorithm}
\usepackage{algpseudocode}

\def\BibTeX{{\rm B\kern-.05em{\sc i\kern-.025em b}\kern-.08em
    T\kern-.1667em\lower.7ex\hbox{E}\kern-.125emX}}
\usepackage[belowskip=-15pt,aboveskip=0pt]{caption}

\begin{document}

\title{Condition monitoring and anomaly detection in cyber-physical systems\\

}

\author{\IEEEauthorblockN{\textsuperscript{} William Marfo}
\IEEEauthorblockA{\textit{Department of Computer Science} \\
\textit{University of Texas at El Paso}\\
El Paso, USA \\
wmarfo@miners.utep.edu}
\and
\IEEEauthorblockN{\textsuperscript{} Deepak K. Tosh}
\IEEEauthorblockA{\textit{Department of Computer Science} \\
\textit{University of Texas at El Paso}\\
El Paso, USA \\
dktosh@utep.edu}

\and
\IEEEauthorblockN{\textsuperscript{} Shirley V. Moore}
\IEEEauthorblockA{\textit{Department of Computer Science} \\
\textit{University of Texas at El Paso}\\
El Paso, USA \\
svmoore@utep.edu}

}

\maketitle

\begin{abstract}
The modern industrial environment is equipping myriads of smart manufacturing machines where the state of each device can be monitored continuously. Such monitoring can help identify possible future failures and develop a cost-effective maintenance plan. However, it is a daunting task to perform early detection with low false positives and negatives from the huge volume of collected data. This requires developing a holistic machine learning framework to address the issues in condition monitoring of high priority components and develop efficient techniques to detect anomalies that can detect and possibly localize the faulty components. This paper presents a comparative analysis of recent machine learning approaches for robust, cost-effective anomaly detection in cyber-physical systems. While detection has been extensively studied, very few researchers have analyzed the localization of the anomalies. We show that supervised learning outperforms unsupervised algorithms. For supervised cases, we achieve near-perfect accuracy of 98\% (specifically for tree-based algorithms). In contrast, the best-case accuracy in the unsupervised cases was 63\%—the area under the receiver operating characteristic curve (AUC) exhibits similar outcomes as an additional metric.

\end{abstract}

\begin{IEEEkeywords}
Machine learning, condition monitoring, anomaly detection, cyber-physical systems
\end{IEEEkeywords}

\section{Introduction}
\label{intro}

Anomaly detection in industrial component health monitoring in cyber-physical systems (CPS) has become more crucial in recent times than at any point in history, with ubiquitous internet of things (IoT) integration in everyday lives and autonomous systems in manufacturing processes and machines. Faults are inevitable in modern machines since all CPS components have a lifespan. With estimated millions of machine failures worldwide and the increasing cost of repairs and financial losses, it is vital to perform early detection and save costs. It is necessary to ensure low false alarms while achieving correct localization. As such, anomaly detection has led many researchers and engineers from energy, military, biomedical, and others to study the application of machine learning (ML) approaches to CPS \cite{ml-survey}. However, compared to simply detection/classification, very few works have addressed localization in this context since fault localization is predominantly studied in software-based systems only. Localization is identifying the exact locations of faults in a system, and it serves as one of our motivations.

The first goal of our work is to demonstrate a complete picture of the application and effectiveness of machine learning (ML) approaches for both the detection and localization of faults in CPS. These approaches have been applied in anomaly detection in cyber-physical systems, resulting in effective and precise outcomes. Using ML approaches can increase the comprehension of condition monitoring as operators need proper validation for these methods for routine monitoring and anomaly detection in cyber-physical systems. As a result, cyber-physical systems require maintenance and security. As a complex system made up of physical and software components, a CPS must be able to detect anomalies since it often operates autonomously in an unpredictable environment. However, detecting anomalies is a challenging task due to the ever-changing nature and lack of a precise model for a CPS. A typical anomaly detection method does not work directly with these challenges due to the growing volume of data and the need for domain-specific knowledge \cite{b26}.

Secondly, most research in literature use ML approaches to detect, classify, and identify anomalies in mid-size to enormous data sets \cite{b9, b28}. Researchers have only recently attempted to utilize machine learning techniques to predict anomalies and localize them. Detecting the CPS anomalies within time constraints is an emerging and significant topic of discussion. We address these limitations in our work.

Another observation of ours lies in the nature of the data that are typical of anomalies of CPS. Naturally, such data sets are imbalanced where the majority of the faulty data points class overwhelms the minority class of faulty points. Although it has been ventured in literature in general contexts, very few focused on precise pre-processing of data imbalance in CPS. We aim to fill this void.

\subsection{Risks and Challenges}
We seek to address the following \textbf{risks} and \textbf{challenges} of anomaly detection in CPS:
both high false positives and negatives, low accuracy, and unbalanced data sets. The challenge of unbalanced data sets arises when the target variable has more observations in a particular class than others, creating more instances of high cases of false positives in most classes.

\textbf{Limitations of unsupervised models:}  It is reported that most unsupervised methods used in condition monitoring in cyber-physical systems are cumbersome, labor-intensive, and ineffective \cite{unsuper}. Extant research also shows that a model like ordering points to identify cluster structure (OPTICS) \cite{b28} is time-consuming and expensive to implement. Such models apply classification techniques that use irrelevant and poorly distributed occurrence data which may not be accurate \cite{DBSCAN-disadvantage}.


\noindent Therefore, we aim to address the following critical research questions in this paper.

\begin{itemize}
  
  \item  How much improvement is achieved when we already have labeled data compared to assigning labels using unsupervised techniques? 
  
  \item Using the set of unsupervised ML algorithms (K-Means clustering, Agglomerative Clustering, OPTICS) and supervised ML algorithms (Logistic Regression, Random Forest, Decision Tree Classifier, AdaBoost Classifier) - predict if the bearings \cite{b32} will produce faults (defined by thresholding quantization error) and in which bearing will it occur (localization)?

  \item Which ML technique, in general, is best for the given data set based on standard performance metrics? 

  \item Which algorithm most reliably performs well in pinpointing anomalous sensors while eliminating false positives and negatives (better localization)? 

\end{itemize}

Section II of the paper discusses related research in cyber-physical condition monitoring and anomaly detection. Section III discusses our methodologies, presenting a brief background of machine learning techniques. Section IV introduces and discusses our data set; Section V spells out the discussion of the theory and background. Section VI presents the results, and the conclusion of this paper is discussed in Section VII.

\section{Related work}
\label{rw}

This section briefly outlines the most recent and widely used techniques in detecting anomalies and monitoring health in cyber-physical systems.
Researchers in \cite{b2} used the self-organizing maps(SOM) machine learning technique to detect anomalies in the bearings data. The authors adopted only SOM in predicting anomalies, where they implemented anomaly localization, error analysis, and degradation.
Researchers in \cite{b4} used artificial neural networks (part of the broader neural networks algorithms family) to create the machine learning model under discussion here. A neural network model was used after testing and training. The numerical analysis and statistics show that this model successfully detects anomalies on the tested photovoltaic system with an accuracy of greater than 90 percent, more than an efficient result \cite{b4}. For photovoltaic systems, anything above 90\% is considered acceptable in photovoltaic systems. 

With the right dataset, one can get the best precision for the ML techniques used in predictive maintenance. The authors in \cite{b1} introduced a study on several past methods applied for predictive maintenance. The dataset used had about 9397 observations where they utilized the expectation framework to anticipate the network failures.
The research primarily centered around the framework level of faults. Moreover, in \cite{b5}, the researchers used a hybrid approach for fusing physics and failure prediction as they could predict the turbine blade failure with accuracy. The authors in \cite{b6} utilized machine learning techniques to anticipate the remaining time to failure (RTTF). 

Furthermore, researchers in \cite{b9} discussed clustering models for machine degradation prediction. They introduced an ensemble model that used solo time-series information to detect anomalies. They then grouped the data and ensured they had good names using supervised learning. However, most of these papers did not address the issue of localization, where specific signals or sensors are pointed out after anomalies are detected since modern cyber-physical systems have large amounts of data, including sensors and control signals. The work we present addresses both classification and localization in fault detection/localization. Moreover, we present a comprehensive comparative analysis between supervised and unsupervised ML approaches used in condition monitoring.

\begin{figure}[h]
\centerline{\includegraphics[width=0.39\textwidth]{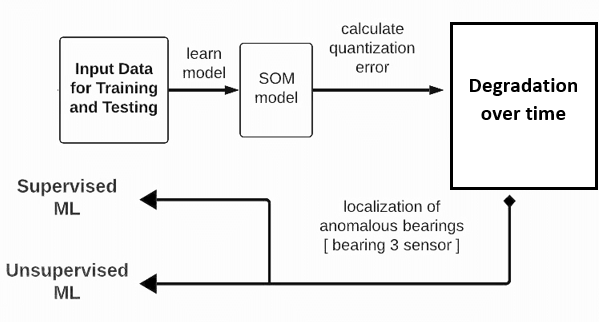}}
\caption{Anomaly detection and localization flow}
\label{fig}
\end{figure}

\section{Methodology}
\label{methodology}

This section briefly describes the candidate machine learning models that effectively detect anomalies from large datasets. The models used include self-organizing map (SOM), random forest, decision tree, logistic regression, KNeighbors, AdaBoost, OPTICS (ordering points to identify clusters structure), Agglomerative and K-means clustering. Furthermore, our approach is novel because we predict the fault or failure time and localize the component, in this case, which of the four bearings will constitute a fault. As such, we could train supervised models in addition to unsupervised models, and the whole procedure is illustrated in Fig. 1. 
 \cite{b6}.
 
 The clustering techniques were configured using Google Colab. OPTICS was based on the confusion matrix. The agglomerative clustering sees the individual bearing dataset as individual clusters. The data's observations are first combined based on similar characteristics until there are no more similar observations to cluster. Then at this stage, we obtain the final clusters. To know if a data point is misplaced here, we use the Cophenetic Correlation Coefficient (CPCC), which examines how the dendrogram perfectly reflects the original pairwise characteristics of our distance measure. A strong correlation between the cophenetic distance and the original distance measure indicates a valid cluster\cite{b6}. Our CPCC value was  0.9333, indicating a solid measure. The cophenetic distance matrix \cite{b31} measures within-group dissimilarity (heights) in a dendrogram. 

The K-means clustering has cluster specification (k), where the centroid selection selects k imaginary points (centroids). The cluster assignment assigns data points to the closest centroid to form k clusters. The centroid update chooses a new centroid to update a cluster. The convergence achievement test means the addition of members does not improve the clusters. To know if a data point is misplaced here, we use the Silhouette metric, which evaluates how well each object lies within its cluster. A higher Silhouette metric gives a good metric and vice versa. We got 0.369, indicating a low Silhouette metric as other clustering methods are more efficient than K-means.

\textbf{Accuracy metrics}:
In our research, our primary focus is the accurate prediction of the fault, so the best measures, in this case, are other metrics such as F1 score, recall, and area under the receiver operating characteristic curve (AUC)  since accuracy alone is not enough. Recall is the ability of an ML model to find all essential cases in the data. In contrast, precision, on the other hand, is the ability to point out only the essential data points. As precision increases, recall decreases and vice versa.

\subsection{Random Forest}
Random forest is an ML approach that requires little pre-processing for anomaly detection compared to other techniques \cite{b6}. The ensemble of trees is constructed from a training data set and internally validated to predict the response based on the predictors for future observations.  
A unit of the decision tree is not robust enough. However, random forest runs numerous decision trees and sums their results to produce a robust and reliable model for detecting anomalies. Many researchers avoid the random forest model because anomaly detection is complex when used, and the localization also consumes too much time. 
 
\subsection{Self-Organizing Maps}
The self-organizing map (SOM) is a neural network associated with vector quantization and visualization. The idea of the SOM algorithm grew out of early neural network models, especially models of associative memory and adaptive learning \cite{b5}. The predecessors in that research direction include the spatially ordered line detectors of \cite{b5} and the neural field model of \cite{b29}. However, the self-organizing power of these early models was relatively weak. The data-driven condition monitoring approach presented in this paper learns models using data where the system is still in its 'new' condition where no degradation has occurred.

In the SOM, we map the input data to the neurons where each neuron is a weight vector of unique dimensionality. We use a random batch training approach and sample the weight vectors and their initial values from the training data to provide various starting stages to the training process. Training takes place over a chosen amount of epochs, and the samples from the training data are fed to the algorithm in one epoch. The best matching unit (BMU) is computed for each input sample from the training data by finding the neuron with the smallest distance to the sample. The BMU and its neighboring neurons, allocated through the neighborhood radius, are moved towards the input sample. Convergence occurs when both neighborhood size and shift strength decline over time.

Eventually, every neuron of the SOM represents a section of the training data. Regions in the training input space with a few examples are represented by a few neurons of the SOM, while a more significant number of neurons represents dense regions. Typically, the quantity of neurons selected is smaller than the number of training data samples, decreasing the training data to the most significant samples.

\textbf{Anomaly detection:}
The SOM is used in anomaly detection by calculating the quantization error. All errors below a threshold are considered normal, while those above are anomalous. This concept of quantization error approaches is already used in network monitoring and anomaly detection of industrial processes. The previous works failed to perform anomaly localization and only used the quantization error in detecting degradation.
The quantization errors for data that are not anomalous are usually positive integers. A threshold is required for the quantization error. It is very prudent to select the threshold manually. However, it is usually unfeasible for practical applications. It is very convenient to estimate the threshold from the data. The training data's quantization error can be considered a probability distribution. The threshold for anomaly detection can be derived from the quantiles with the help of the probability distribution. When labels are present, the quantiles can be used to fine-tune the score of the anomaly detection. The quantile can adjust the optimization of the outcome of the anomaly detection.
\newline
The threshold value for the quantization errors to classify a sample input as an anomaly or not was selected based on quantile values. Quantile returns estimates of underlying distribution based on order statistics from input values for selected probabilities. To get the optimal threshold value, the model searches through a range of probability values (from 0 to 1) for the training data. In this study, the optimal value occurred at 17.82\%. Thus, sample inputs with quantization errors above the 18th percentile are considered an anomaly, as indicated by the orange line in Fig.3.

\smallskip
\textbf{Anomaly localization:}
After the anomaly detection, we calculate the signal or sensor where the anomaly occurred. The anomalous observation is fed through a reverse model to get the expected values for the signals. We use the deviations from the expected values to identify the sensor related to the anomaly. Once the degradation is detected, we retrieve the BMU by mapping the input sample to the SOM. In the SOM, each neuron's weight vector has the same dimension as the input data and contains its corresponding signal value. After the distance of the weight vector to each signal is calculated, the resulting distances and their signals are sorted by their distances in descending order. There are usually many signals in real-world cyber-physical systems, so it is essential to reduce the number of signals displayed. Therefore we only display the first\emph{ n} signals giving operators a starting point to locate the likely faults in the systems to restore the regular working order of the systems. Only the signal with the most significant deviation (n=1) was considered in our paper.

Degradation of the system leads to a deviation from the detected condition over time. We expect the deviation between the models to increase over the system's lifetime. The data are learned, and the error threshold is estimated accordingly, where we then map the new data to the SOM and calculate the deviation. The localization also provides more information about the anomalies, so the specific part of the system can be pinpointed and restored to full functionality. 

\smallskip
\section{Data background}
\label{data}
We present a brief background of the dataset generated by the NSF I/UCR Center for Intelligent Maintenance Systems with support from Rexnord Corp. in Milwaukee, WI. 
In the setup, four bearings have been hooked up on a shaft. The rotation pace was maintained constant at 2000 RPM through an AC motor. A 0.45 kg radial load is applied onto the shaft and bearing through a spring mechanism. Rexnord ZA-2115 double row bearings were hooked up at the shaft, as shown in Fig. 2. PCB 353B33 excessive Sensitivity Quartz ICP accelerometers were mounted on the bearing housing. The general placement of the sensors is also shown in Fig. 2. All faults occurred after exceeding the designed life of the bearing, which is greater than 100 million revolutions.

Each data set describes an experiment to test failure, incorporating files that are 1-second vibration signal snapshots recorded at specific durations. The data set has 984 files with 20480 observations each. Every file consists of 20,480 points, with the sampling rate set at 20 kHz. The second set of data was used in our experiment for this research.

\begin{figure}[h]
\centerline{\includegraphics[width=0.43\textwidth]{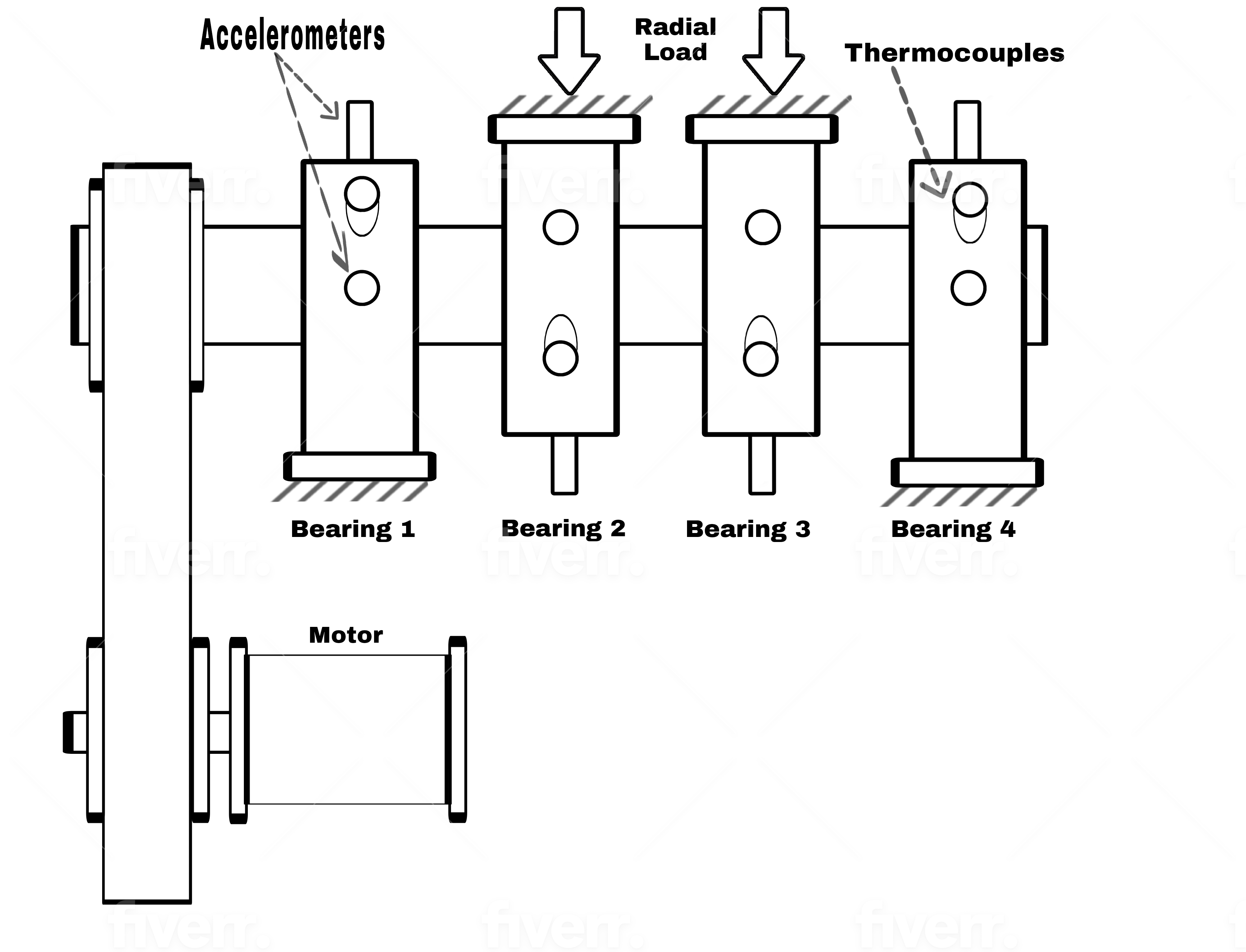}}
\caption{Bearing test rig set up and placement of sensor.}
\smallskip
\label{fig}
\end{figure}

\section{Discussion of theory and background}
First, after training the SOM to represent the 'new' condition of the system, the quantization error is used to derive the system's overall condition. We then use the algorithm for
anomaly localization to pinpoint which sensor is responsible for the degradation of the
condition. Data set number two is used in our experiment, and we select the first 10 percent of the data to represent the new condition. The SOM is trained on our chosen data with 1,843,200 samples. 
The SOM is set to 50x50 neurons by size, and training takes over 100 epochs. The data are carefully normalized to a range between 1 and 0. At the computation efforts during the training and search for BMU, larger SOM represents the training data in more detail than a smaller one.
The quantization error of the training data is calculated and set to 0.9999 for the estimation of the training error deviations less than the estimated threshold will be considered normal, as indicated in Fig. 3.

\begin{figure}[h]
\centerline{\includegraphics[width=0.34\textwidth]{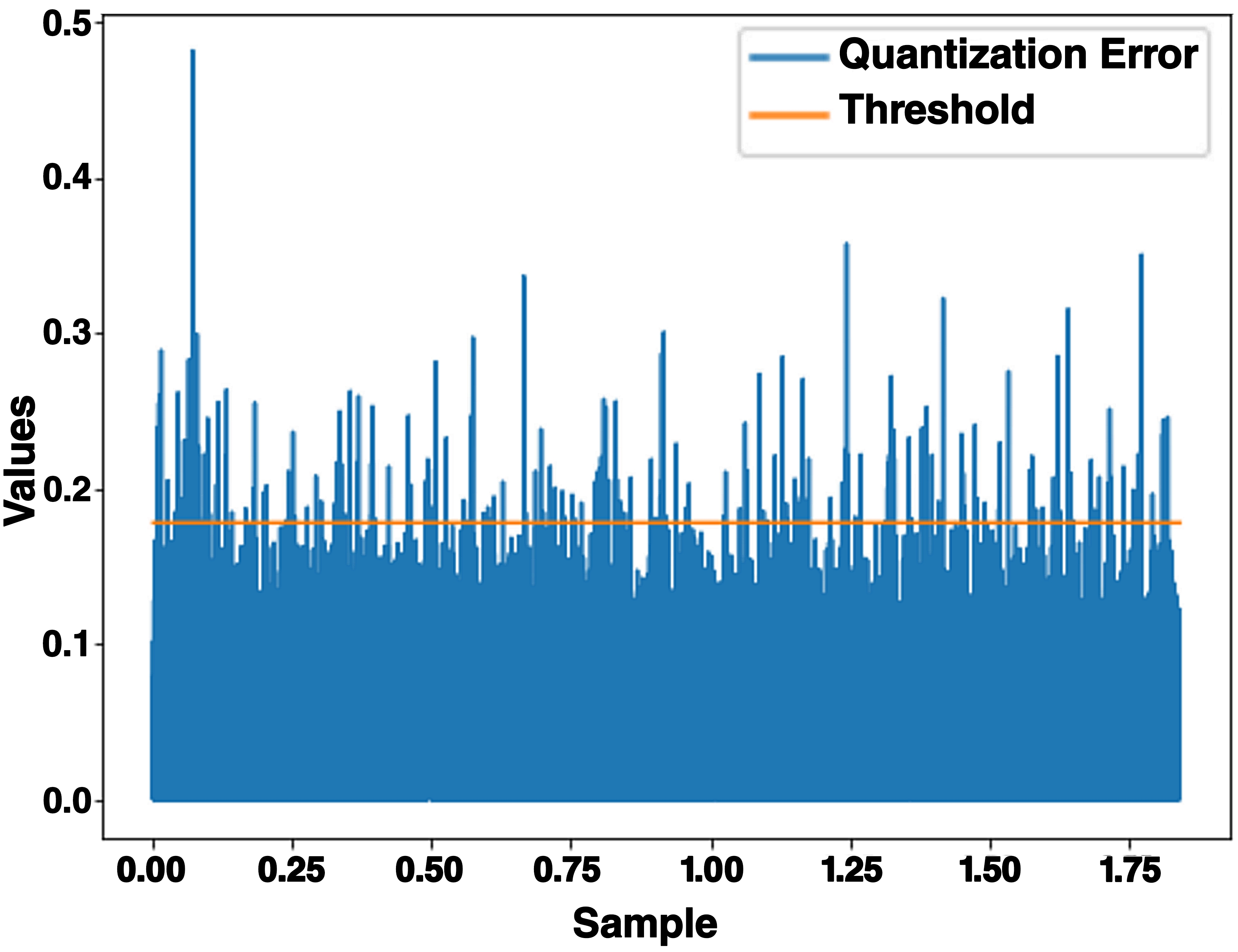}}
\caption{Estimated threshold of the training data for the second bearing data set.}\smallskip
\label{fig}
\end{figure}

We next compare the complete data set to the learned SOM and then calculate the quantization error for every sample. 
 Right after February 17, we noticed a spike in the degradation in bearing 1 (Figure 4). It is noted that the deviation on February 17, which started around 7:42 in the morning, is caused by bearing 1, which translates to the first column in the data, specifically its vibration sensor. This continues until total degradation occurs after February 19, when the experiment is stopped. Our model detected the bearing 1 sensor failure two days before its failure. Operators can use this to prepare and react to system failures before complete failure occurs. In Fig. 4, the observations in each file are localized, and their occurrences are counted and plotted.

In normal conditions, we expect an equal distribution across all signals. The failed sensors are drawn more often than the other sensors. Based on the plot, all bearings are drawn equally at the start, but once failure begins, bearing 1 is drawn more frequently than the others.

\begin{figure}[h]
\centerline{\includegraphics[width=0.38\textwidth]{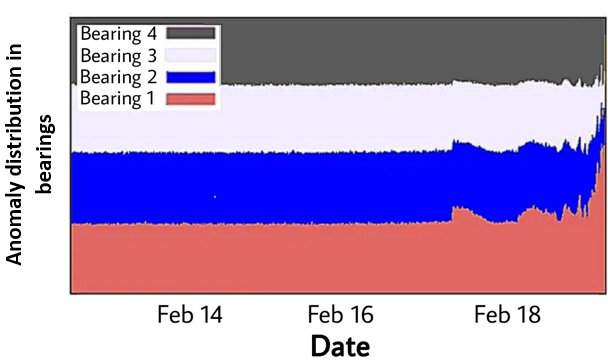}}
\caption{Anomaly localization in the second bearing data set.}
\label{fig}
\end{figure}

\section{Results}
The contribution of this paper is to predict and detect anomalies in the bearings data while generating predictive alerts. We evaluated our self-organizing maps model using the second set of bearing data to locate specific anomalies in the sensors after computing the quantization error, which addresses gaps in most existing research. We also compared the performance metrics of the techniques utilized by combining supervised and unsupervised techniques in terms of their F1 score, precision, and recall. Our research questions showed a considerable improvement in the results when labeled data was used. Secondly, the supervised tree-based models like a decision tree and random forest performed best.


\subsection{Summary of performance of  Ml algorithms }

After training 80\% and testing 20\%, we summarize the models' performance after detecting the anomalies with the SOM model. In Fig. 5, the accuracy of the clustering techniques used in the research is plotted in a bar graph. The agglomerative technique outperforms the other clustering algorithms.

\begin{figure}[h]
\centerline{\includegraphics[width=0.34\textwidth]{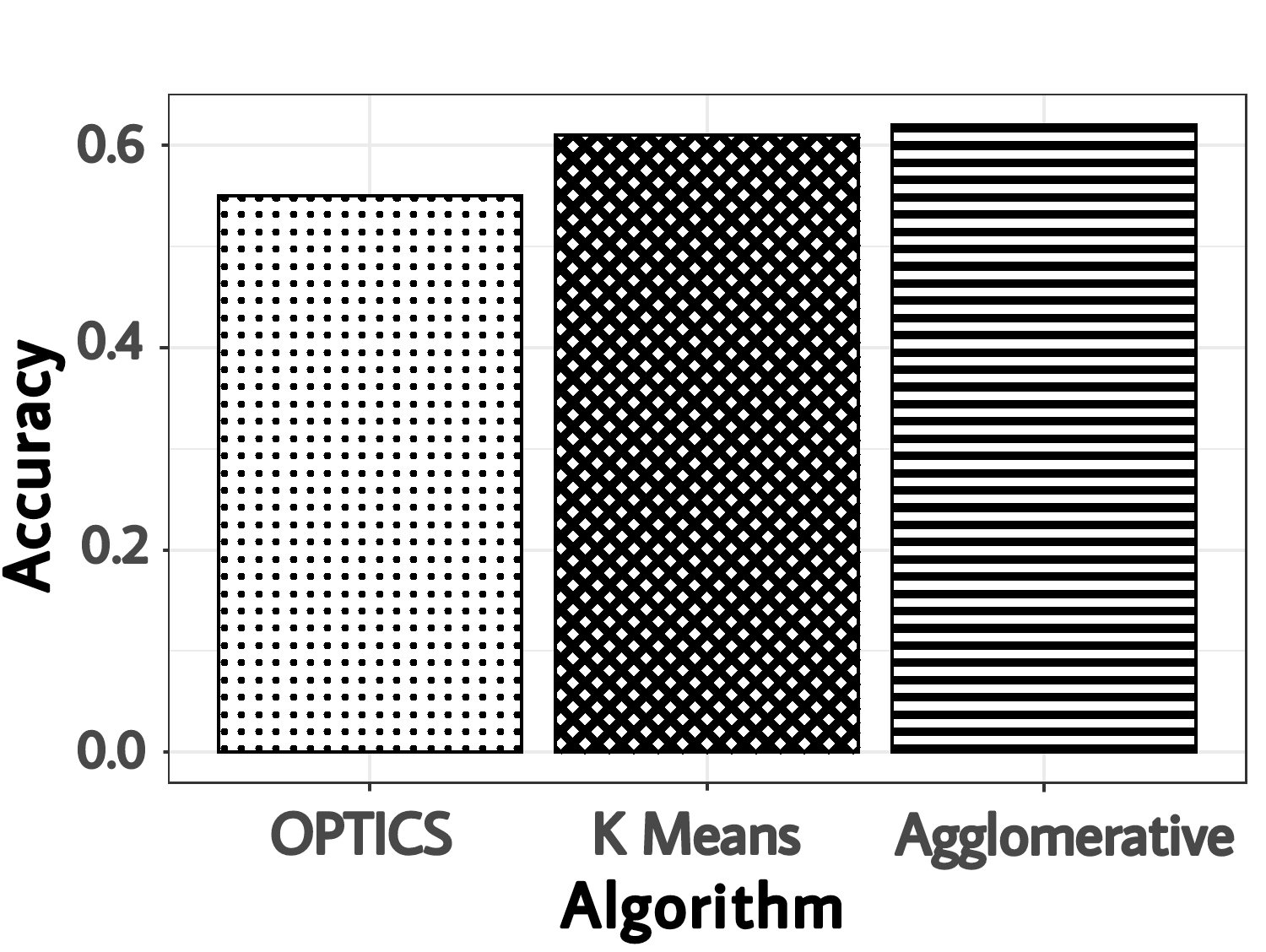}}
\caption{Accuracy of clustering techniques.}
\label{fig}
\end{figure}

\begin{figure}[h]
\centerline{\includegraphics[width=0.33\textwidth]{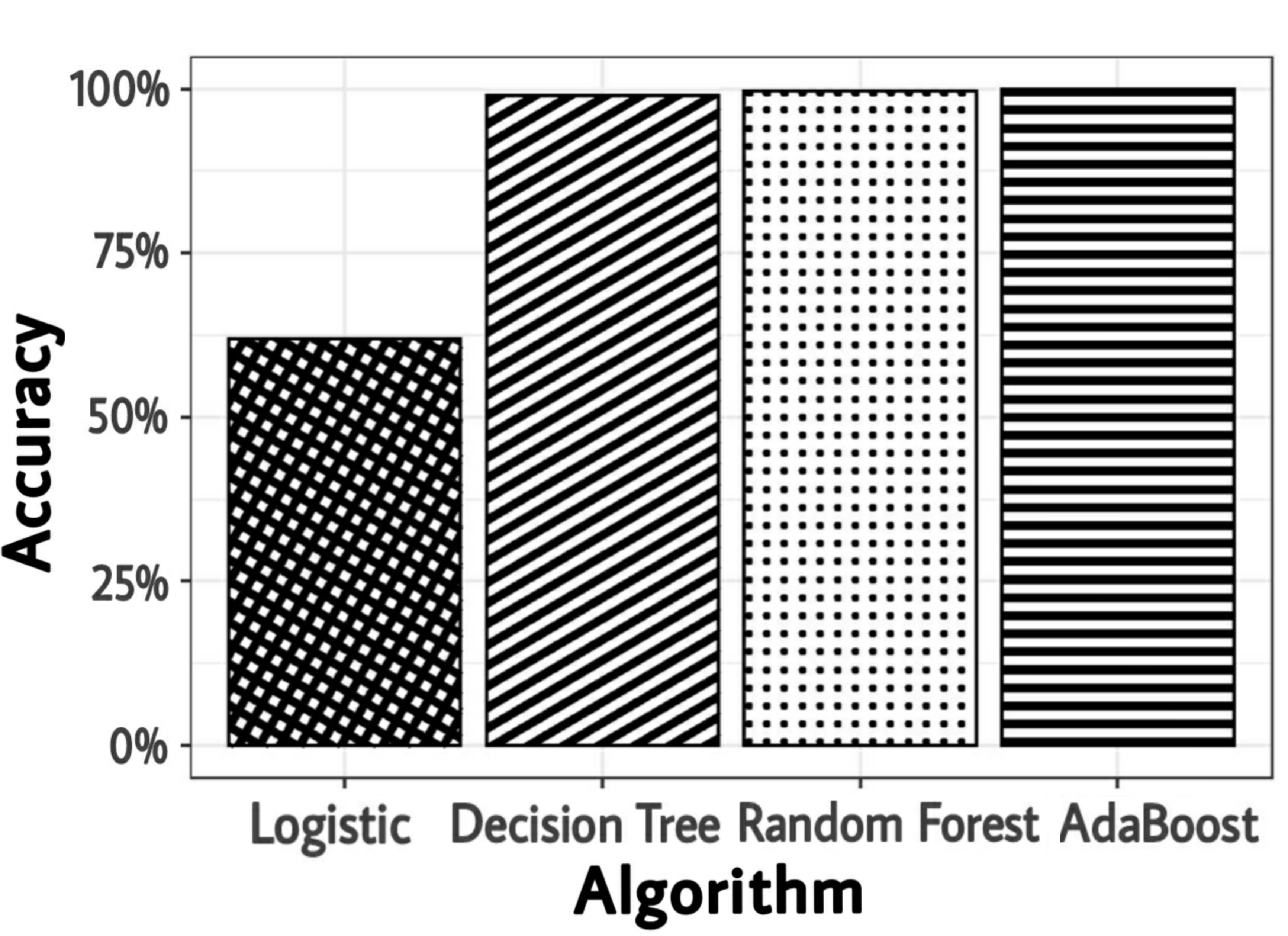}}
\caption{Accuracy of classification techniques.}
\label{fig}
\end{figure}

\subsection{Accuracy and Error Metrics} \label{metrics2} Evaluating a machine learning algorithm is an essential part of any research. Model accuracy can be used to measure model performance, but it is not enough to assess its effectiveness. In Fig. 7, we present the accuracy, precision, recall, and F1 score of the unsupervised algorithms used. It is noticeable that the unsupervised algorithms struggled to obtain a good recall score. This is because most of the predictions were unexpectedly on the wrong side.

\begin{figure}[htbp]
\centerline{\includegraphics[width=0.34\textwidth]{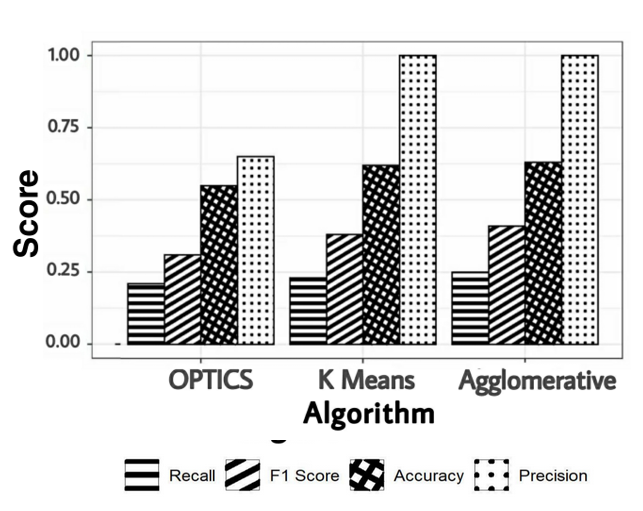}}
\caption{Unsupervised algorithms performance measures for bearing data set.}\smallskip

\label{fig}
\end{figure}

\begin{figure}[h]
\centerline{\includegraphics[width=0.34\textwidth]{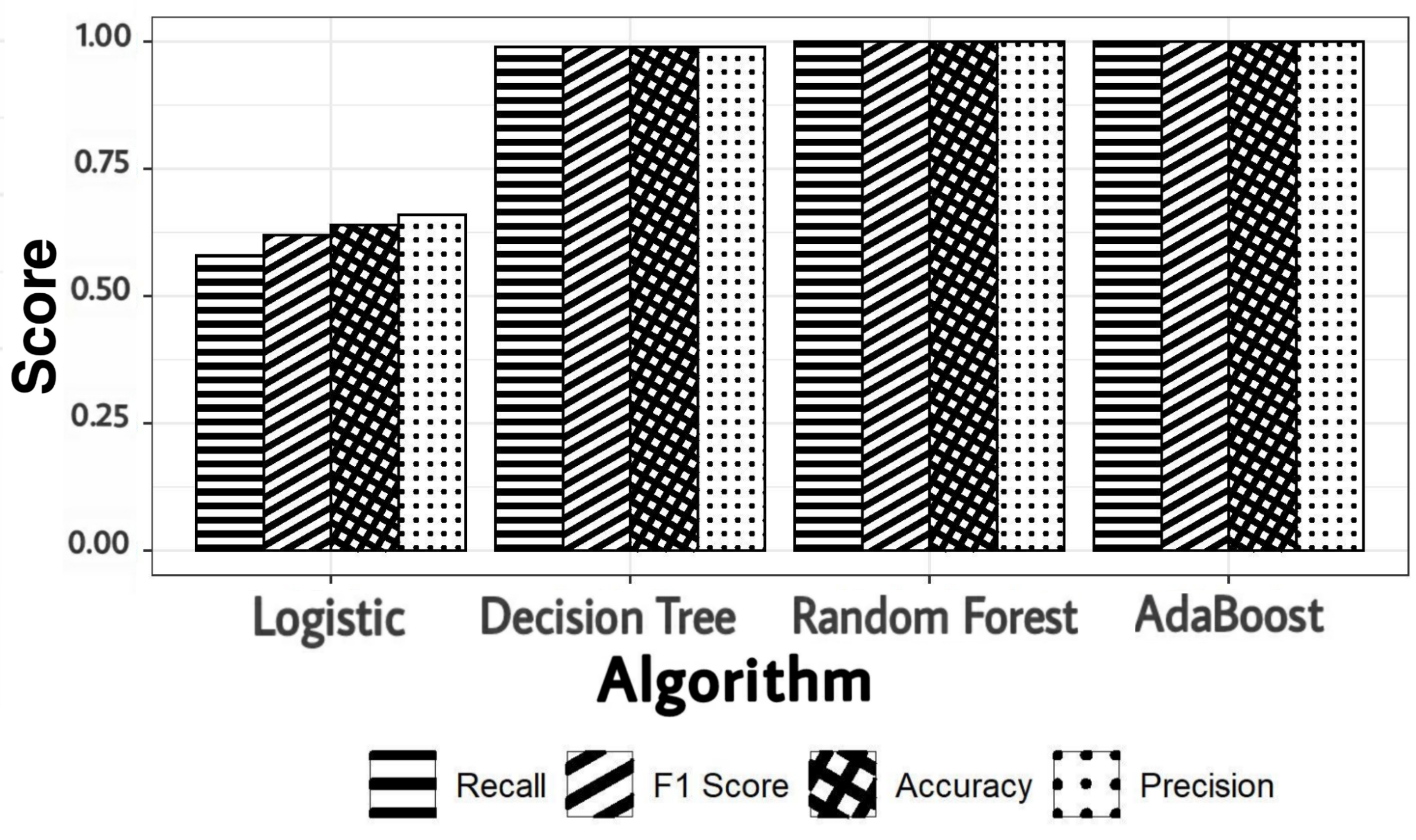}}
\caption{Supervised algorithms performance measures for bearing data set.}\smallskip
\label{fig}
\end{figure}

The accuracy of the classification techniques used for the fault prediction is presented in Fig. 6, and almost all the algorithms have better accuracy than in Fig. 5. AdaBoost has the highest accuracy together with random forest, whereas logistic regression is near. The higher performance of Adaboost and Random Forest is likely because of them being ensemble ML models, whereas logistic regression is not. Thus, the rationale lies in the robustness against the bias of the ensemble models. We compute the performance metrics in Fig. 8 and Fig. 7 for supervised and unsupervised techniques, respectively. The results look similar in Fig. 8, and they give us a positive indication that our predictions are correct.

\begin{table}[h]
\vspace*{1mm}
\smallskip
\caption{Performance evaluation}
\label{tab:Classification Analysis}
\begin{tabular}{|l|l|l|l|l|}
\hline
\textbf{Algorithms}       & \textbf{F1 Score} & \textbf{Precision} & \textbf{Recall} & \textbf{Accuracy} \\ \hline
K Means Clustering       & 0.38              & 0.99                  & 0.23            & 0.62              \\ \hline
Agglomerative Clustering & 0.41              & 0.99                  & 0.25            & 0.63              \\ \hline
                             
OPTICS                   & 0.31              & 0.65               & 0.21            & 0.55              \\ \hline
Logistic Regression      & 0.62              & 0.66               & 0.58            & 0.64              \\

\hline
Random forest      & 0.99                 & 0.99                  & 0.99               & 0.99                 \\

\hline
Decision Tree Classifier & 0.99              & 0.99               & 0.99            & 0.99              \\

\hline
AdaBoost Classifier      & 0.99                 & 0.99                  & 0.99               & 0.99                 \\ \hline
\end{tabular}
\end{table}

Precision represents how precise/accurate each model is out of those data points that are predicted as faulty and how many of them are actually faulty. A high precision (seen in AdaBoost and Random Forest) is a good measure considering an unbalanced data set and is robust against false positives. (False positives occur when a model mispredicts the positive outcome. An incorrectly predicted negative class is referred to as a false negative). The recall represents how many of the actual faults the models captured through predicting it as faulty. Moreover, AdaBoost and Random Forest have the highest recall, showing the least false negatives. F1-score is a balance of both precision and recall, given as $F1\: score=2\times\frac{precision \times recall}{precision + recall}$ . Thus, it makes sure that high precision or recall alone cannot bias the metric. 

Finally, accuracy is the simple mean of correctness derived from the difference in predictions from the labeled ground truth data. In our performance evaluation in Table 1, the supervised tree-based models like decision trees, AdaBoost, and random forests performed well. This is further plotted as solid evidence on our ROC plot in Fig. 9.\smallskip

\begin{figure}[h]
\vspace*{-4mm}
\centerline{\includegraphics[width=0.53\textwidth]{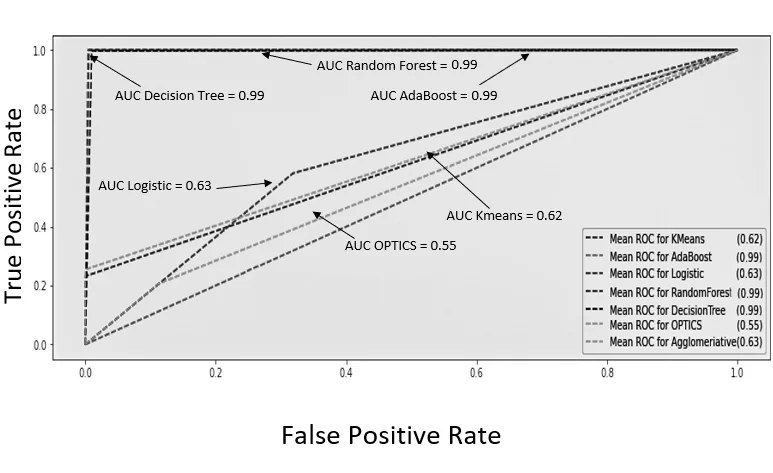}}
\caption{Model evaluation using AUC.}\smallskip \smallskip
\label{fig}
\end{figure}

\textbf{The Receiver Operating Curve (ROC) }helps us evaluate the performance of our models by indicating the relationship between true and false classes. The AUC separates true positive and false-positive rates. Hence the higher the AUC, the better the model's performance at distinguishing between the positive and negative classes. From Fig. 9, it can be seen that the tree-based algorithms have the highest AUC as Decision Tree, Random Forest, and AdaBoost have outstanding results of 0.99, 1, and 1, respectively.

\section{Conclusion and future work }
In this work, we presented a comprehensive and comparative analysis of supervised and unsupervised ML approaches in detecting and localizing anomalies during condition monitoring of a cyber-physical system represented by a large-bearing data set. We also developed and introduced a framework that detects and localizes faults in bearings that is robust to data imbalance. Our research showed that the supervised tree-based models like decision trees and random forests performed best. 

There is still a lot to unearth regarding ML and anomaly detection in cyber-physical systems. As component health deteriorates over time as part of their average lifestyle, it sometimes leads to replacement at fixed intervals as it may occur that they are not faulty. By detecting anomalies with ML, these defects can be predicted promptly. In the future, the estimation and development of complex fault classification rules may also aid in predicting anomalies by supplementing ML approaches. Additionally, increasing the scope of this research and incorporating federated ML may allow researchers to estimate the remaining life of cyber-physical systems by training data in a distributed approach.

\end{document}